\documentclass{midl} 

\usepackage{mwe} 

\usepackage{float}
\usepackage{amsmath, multicol, multirow, booktabs, microtype, balance}
\usepackage[font=small,skip=0.25cm]{caption}

\jmlrproceedings{MIDL}{Medical Imaging with Deep Learning}
\jmlrpages{}
\jmlryear{2023}

\jmlrworkshop{Short Paper -- MIDL 2023 submission}

\title[Forward-Forward Contrastive Learning]{Forward-Forward Contrastive Learning}

 \midlauthor{\Name{Md. Atik Ahamed} \Email{atikahamed@uky.edu}\\
  \Name{Jin Chen} \Email{chen.jin@uky.edu}\\
  \Name{Abdullah-Al-Zubaer Imran} \Email{aimran@uky.edu}\\
  \addr University of Kentucky, Lexington, KY, USA}

\begin{document}

\maketitle

\begin{abstract}
Medical image classification is one of the most important tasks for computer-aided diagnosis. Deep learning models, particularly convolutional neural networks, have been successfully used for disease classification from medical images, facilitated by automated feature learning. However, the diverse imaging modalities and clinical pathology make it challenging to construct generalized and robust classifications. Towards improving the model performance, we propose a novel pretraining approach, namely \textbf{Forward Forward Contrastive Learning (FFCL)}, which leverages the Forward-Forward Algorithm in a contrastive learning framework--both locally and globally. Our experimental results on the chest X-ray dataset indicate that the proposed FFCL achieves superior performance (\textbf{3.69\%} accuracy over ImageNet pretrained ResNet-18) over existing pretraining models in the pneumonia classification task. Moreover, extensive ablation experiments support the particular local and global contrastive pretraining design in FFCL.
\end{abstract}


\section{Introduction}

The imperative for automated disease diagnosis is contingent upon the accurate classification of medical images. In this context, deep convolutional neural networks (CNNs) have proven to be remarkably effective in performing medical image classification tasks, thereby significantly contributing to the advancement of the automated diagnosis process. Nevertheless, CNNs exhibit limitations in terms of generalizability and the ability to capture fine details within input images. To address these limitations, we propose a multistage pretraining method with back-propagation, which effectively improves the  performance of medical image classification and enhances model generalizability. 
According to the forward-forward algorithm (FFA) in~\citet{hinton2022forward}, there is no convincing evidence to suggest that our brain stores gradients and undergoes learning via a back-propagation mechanism. Moreover, the majority of contemporary disease classification tasks are carried out by training state-of-the-art deep learning models using the back-propagation approach. These models were typically trained from scratch using medical images or fine-tuned based on the ImageNet-pretrained models~\citep{deng2009imagenet}. However, ImageNet does not well represent the characteristics of images within the medical imaging domain, resulting in suboptimal model generalizability. 
In this project, we leverage supervised contrastive learning~\citep{khosla2020supervised} as a pretraining strategy, instead of directly utilizing back-propagation with weights pretrained on the ImageNet dataset. Existing studies demonstrate contrastive learning performed by only taking the final output of the model, which does not capture the fine image details. In this work, we propose a multistage pretraining strategy, called FFCL, before performing back-propagation. We pretrain our backbone model in a supervised contrastive representation learning manner locally for each layer, and globally for the target model to capture fine details of the input image. Our pretraining strategy is inspired by FFA. However, our solution does not require manual fine-tuning of any thresholds, which helps the model to capture local information automatically and reduce manual effort. To the best of our knowledge, this is the first work leveraging contrastive learning in a forward-forward mechanism in medical imaging. Notably, the proposed FFCL can be extended further to train a model end-to-end without requiring any manual intervention in between.

 \begin{figure}
    \centering
    \includegraphics[width=\linewidth]{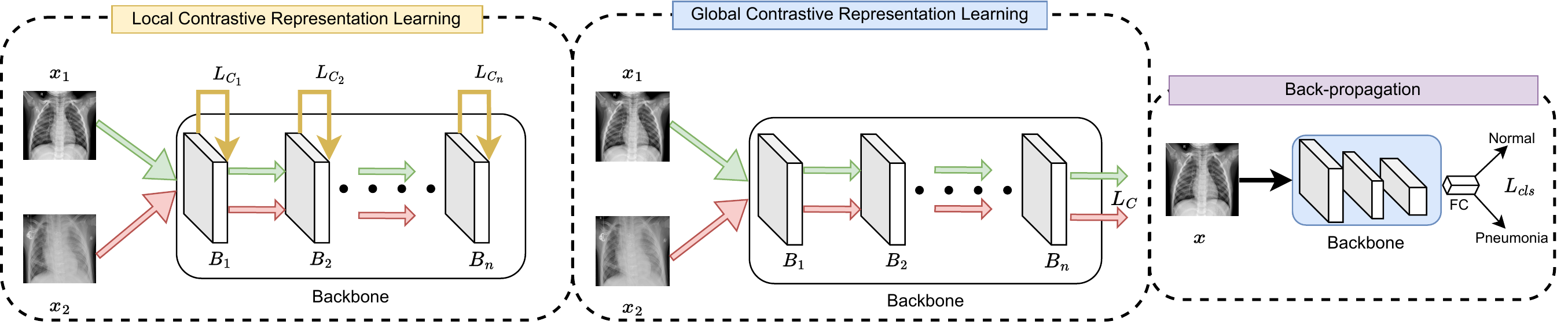}
    \caption{Proposed FFCL model: $L_{C_i}$ and \ $L_C$ represent local and global contrastive loss, respectively. $B_i, i=[1,n]$ represents a block of the target model.}
    \label{fig:method}
\end{figure}

\section{Methods}
FFCL comprises two stages of pretraining before performing regular back-propagation for the downstream classification tasks. Fig.~\ref{fig:method} shows the FFCL framework, which comprises two pretraining stages (local and global contrastive representation learning) and the final back-propagation-based image classification. In the first stage, we perform contrastive learning locally, based on the modified FFA~\citet{hinton2022forward}. Unlike FFA, we leverage supervised contrastive learning for local updates at each layer of the target model without requiring tuning any thresholds. In the first stage, FFCL randomly takes two images $x_1, x_2 \in X_{train}$ sampled from the train set ($X_{train}$) and provides embeddings $E_{{x_1}_i}, E_{{x_2}_i}$ from each block $B_i$ followed by ReLU~\citep{agarap2018deep} activation layer. We perform local updates utilizing cosine embedding loss $L_{C_i}$ for each of the blocks ($B_i$s). After performing local updates for each block, the first stage pretrained model is used for performing the global contrastive learning in the second stage. As in local contrastive learning, this global contrastive learning takes two random images as input and maps to the final embedding space ($E_{x_1}, E_{x_2}$). The same cosine embedding loss is used for global contrastive learning (Eq. \ref{eq:loss_contrastive}). In the third stage, the latest pretrained model is leveraged for performing the actual downstream classification task with regular back-propagation. For baselines and FFCL, we use the binary-cross-entropy loss for the downstream classification task. FFCL is elegant in the sense all the training stages are performed in a fully automated manner without requiring any hyper-parameter tuning in between.
\begin{equation}
\operatorname{L_c}(E_{x_1},E_{x_2},C_{x_1},C_{x_2})= \begin{cases}1-\frac{E_{x_1} \cdot E_{x_2}}{\lVert E_{x_1} \rVert_2 \lVert E_{x_2} \rVert_2}, & \text { if } C_{x_1}=C_{x_2} \\ \max \left(0, \frac{E_{x_1} \cdot E_{x_2}}{\lVert E_{x_1} \rVert_2 \lVert E_{x_2} \rVert_2}\right), & \text { if }  C_{x_1}\neq C_{x_2}\end{cases}
\label{eq:loss_contrastive}
\end{equation}
where,  $(E_{x_1}, C_{x_1}) \ \&\ (E_{x_2}, C_{x_2})$ denote the embeddings and class labels of the corresponding input images $x_1\ \& \ x_2$.

\section{Experiments and Results}
We perform binary classification (pneumonia vs. normal) from chest X-ray images employing the ResNet-18 and ResNet-34 backbones~\citep{he2016deep}. We evaluated the proposed method using the pediatric chest X-ray dataset~\citet{kermany10labeled}. The dataset contains a total of 5856 chest X-ray images split into train (5232) and test (624). 
For our experiments, we further separated 262 images from the train set as the validation set which was used for selecting the best models. 
All the images were resized to $224\times224\times3$ and 0--1 normalized (training from scratch) and ImageNet normalized (training from ImageNet weight) conducting experiments with the backbone. The models were trained with a Cosine Annealing scheduler using the Adam optimizer (initial learning rate of 0.0001). We used a batch size of 10 and trained for 100 epochs for all the models for the downstream classification task.

Table~\ref{tab:performance} compares the classification performance of our proposed approach against the baselines. For simplicity, we refer the regular backpropagation as RBP. Except accuracy and ROC-AUC, other reported metrics have been \textit{macro} averaged. Table~\ref{tab:performance} demonstrates superior performance over the baseline RBP with an improvement of 3.69\% in terms of accuracy. Moreover, our proposed FFCL method outperforms state-of-the-art pneumonia classification methods: AUC of 77\% \citet{jaiswal2021scalp}, AUC of 78.4\% \citet{seyyed2020chexclusion} and AUC 75\%~\citet{liu2019align}.

\begin{table}
\centering
\caption{Pneumonia classification performance with the ResNet-18 and ResNet-34 backbone networks. We compare  FFCL against regular backpropagation (RBP), with random and ImageNet pretrained weight initializations. The ablation study is reported on FFCL using ResNet-18.}
\medskip
\label{tab:performance}
\resizebox{\linewidth}{!}{
\begin{tabular}{lc cc cc cc ccccc} 
\toprule
Backbone & \phantom{a} & Approach & \phantom{a} & Contrastive & \phantom{a} & Initialization & \phantom{a}  & Accuracy & F1 & Precision & Recall & AUC\\ 
\midrule
\multirow{9}{*}{ResNet-18} && \multirow{2}{*}{RBP} && -- && ImageNet && 76.76 & 69.84 & 85.94 & 69.10 & 92.90\\
&& && -- && random && 73.08 & 63.14 & 84.95 & 64.10 & 89.15\\
\cmidrule{3-13}
&& \multirow{7}{*}{FFCL} && Local$\rightarrow$Global && ImageNet &&  72.28 & 61.60 & 84.64 & 63.03 & 84.21\\
&& && Local$\rightarrow$Global && random && \textbf{80.45} & \textbf{75.69} & \textbf{87.70} & \textbf{74.02} & \textbf{93.33}\\
&& && Global$\rightarrow$Local && random && 74.04 & 65.38 & 83.52 & 65.64 & 91.23\\
&& && Global only && ImageNet &&  69.71 & 56.38 & 83.68 & 59.62 & 87.88\\
&& && Local only && ImageNet &&  74.36 & 65.52 & 85.45 & 65.81 & 93.09\\
&& && Global only && random &&  76.92 & 70.22 & 85.54 & 69.40 & 92.25\\
&& && Local only && random &&  62.50 & 38.46 & 31.25 & 50.00 & 50.00\\
\midrule
\multirow{2}{*}{ResNet-34} && RBP &&  -- && ImageNet && 78.53 & 72.71 & \bf86.77 & 71.45 & 94.09\\
&& FFCL && Local$\rightarrow$Global && random && \bf78.85 & \bf73.32 & 86.51 & \bf71.97 & \bf94.43\\
\bottomrule
\end{tabular}
}
\end{table}
\section{Conclusion}
We proposed a novel multistage contrastive pretraining strategy (FFCL) to enhance disease detection with state-of-the-art deep learning models. Our proposed FFCL-based pretraining fully exploits the deep learning models' learnability by performing local and global updates. Our extensive experimentation along with innovative ablation study confirmed the superiority of FFCL over regular backpropagation training in performing pneumonia disease classification. Our ongoing efforts include evaluating on larger-scale datasets of varying diseases as well as additional medical image analysis tasks.

\bibliography{midl-samplebibliography}

\begin{thebibliography}{9}
\providecommand{\natexlab}[1]{#1}
\providecommand{\url}[1]{\texttt{#1}}
\expandafter\ifx\csname urlstyle\endcsname\relax
  \providecommand{\doi}[1]{doi: #1}\else
  \providecommand{\doi}{doi: \begingroup \urlstyle{rm}\Url}\fi

\bibitem[Agarap(2018)]{agarap2018deep}
Abien~Fred Agarap.
\newblock Deep learning using rectified linear units (relu).
\newblock \emph{arXiv preprint arXiv:1803.08375}, 2018.

\bibitem[Deng et~al.(2009)Deng, Dong, Socher, Li, Li, and
  Fei-Fei]{deng2009imagenet}
Jia Deng, Wei Dong, Richard Socher, Li-Jia Li, Kai Li, and Li~Fei-Fei.
\newblock Imagenet: A large-scale hierarchical image database.
\newblock In \emph{2009 IEEE conference on computer vision and pattern
  recognition}, pages 248--255. Ieee, 2009.

\bibitem[He et~al.(2016)He, Zhang, Ren, and Sun]{he2016deep}
Kaiming He, Xiangyu Zhang, Shaoqing Ren, and Jian Sun.
\newblock Deep residual learning for image recognition.
\newblock In \emph{Proceedings of the IEEE conference on computer vision and
  pattern recognition}, pages 770--778, 2016.

\bibitem[Hinton(2022)]{hinton2022forward}
Geoffrey Hinton.
\newblock The forward-forward algorithm: Some preliminary investigations.
\newblock \emph{arXiv preprint arXiv:2212.13345}, 2022.

\bibitem[Jaiswal et~al.(2021)Jaiswal, Li, Zander, Han, Rousseau, Peng, and
  Ding]{jaiswal2021scalp}
Ajay Jaiswal, Tianhao Li, Cyprian Zander, Yan Han, Justin~F Rousseau, Yifan
  Peng, and Ying Ding.
\newblock Scalp-supervised contrastive learning for cardiopulmonary disease
  classification and localization in chest x-rays using patient metadata.
\newblock In \emph{2021 IEEE International Conference on Data Mining (ICDM)},
  pages 1132--1137. IEEE, 2021.

\bibitem[Kermany et~al.(2018)Kermany, Zhang, and Goldbaum]{kermany10labeled}
Daniel Kermany, Kang Zhang, and Michael Goldbaum.
\newblock Labeled optical coherence tomography (oct) and chest x-ray images for
  classification (2018).
\newblock \emph{Mendeley Data, v2 https://doi. org/10.17632/rscbjbr9sj
  https://nihcc. app. box. com/v/ChestXray-NIHCC}, 2018.

\bibitem[Khosla et~al.(2020)Khosla, Teterwak, Wang, Sarna, Tian, Isola,
  Maschinot, Liu, and Krishnan]{khosla2020supervised}
Prannay Khosla, Piotr Teterwak, Chen Wang, Aaron Sarna, Yonglong Tian, Phillip
  Isola, Aaron Maschinot, Ce~Liu, and Dilip Krishnan.
\newblock Supervised contrastive learning.
\newblock \emph{Advances in neural information processing systems},
  33:\penalty0 18661--18673, 2020.

\bibitem[Liu et~al.(2019)Liu, Zhao, Fei, Zhang, Wang, and Yu]{liu2019align}
Jingyu Liu, Gangming Zhao, Yu~Fei, Ming Zhang, Yizhou Wang, and Yizhou Yu.
\newblock Align, attend and locate: Chest x-ray diagnosis via contrast induced
  attention network with limited supervision.
\newblock In \emph{Proceedings of the IEEE/CVF International Conference on
  Computer Vision}, pages 10632--10641, 2019.

\bibitem[Seyyed-Kalantari et~al.(2020)Seyyed-Kalantari, Liu, McDermott, Chen,
  and Ghassemi]{seyyed2020chexclusion}
Laleh Seyyed-Kalantari, Guanxiong Liu, Matthew McDermott, Irene~Y Chen, and
  Marzyeh Ghassemi.
\newblock Chexclusion: Fairness gaps in deep chest x-ray classifiers.
\newblock In \emph{BIOCOMPUTING 2021: proceedings of the Pacific symposium},
  pages 232--243. World Scientific, 2020.

\end{thebibliography}

\end{document}